\documentclass[review]{elsarticle}

\usepackage{hyperref}
\usepackage{amsmath}
\usepackage{multirow}
\usepackage{graphicx}
\usepackage{lscape}
\usepackage{adjustbox}
\usepackage{xcolor}
\usepackage{soul}
\usepackage[utf8]{inputenc}
\usepackage{booktabs}

\usepackage{amsmath}
\usepackage{textcomp}  
\usepackage{lipsum}    
\usepackage{subfig}    
\usepackage{hyperref}
\usepackage{multirow}
\usepackage{graphicx}
\usepackage{lscape}
\usepackage{adjustbox}
\usepackage{pdflscape}
\usepackage{everypage}
\usepackage{lipsum}
\usepackage{etoolbox}
\pretocmd{\abstractname}{\newpage}{}{}

\newcommand{\Lpagenumber}{\ifdim\textwidth=\linewidth\else\bgroup
  \dimendef\margin=0 
  \ifodd\value{page}\margin=\oddsidemargin
  \else\margin=\evensidemargin
  \fi
  \raisebox{\dimexpr -\topmargin-\headheight-\headsep-0.5\linewidth}[0pt][0pt]{%
    \rlap{\hspace{\dimexpr \margin+\textheight+\footskip}%
    \llap{\rotatebox{90}{\thepage}}}}%
\egroup\fi}
\AddEverypageHook{\Lpagenumber}
\usepackage{xcolor}

\usepackage{algorithm,algorithmic}
\usepackage{caption}
\newlength\myindent
\newcommand\bindent{%
  \begingroup
  \setlength{\itemindent}{\myindent}
  \addtolength{\algorithmicindent}{\myindent}
}
\newcommand\eindent{\endgroup}

\journal{MDPI, Data}

\biboptions{authoryear}

\begin{document}
\begin{frontmatter}

\title{A Multi-Phase Approach for Product Hierarchy Forecasting in Supply Chain Management: Application to MonarchFx Inc.}

\author[mymainaddress]{Sajjad Taghiyeh\corref{mycorrespondingauthor}}
\cortext[mycorrespondingauthor]{Corresponding author}
\ead{staghiy@ncsu.edu}
\author[mysecondaryaddress]{David C Lengacher}
\author[mysecondaryaddress2]{Amir Hossein Sadeghi}
\author[mysecondaryaddress3]{Amirreza Sahebi Fakhrabad}
\author[mysecondaryaddress4]{Robert B Handfield}

\address[mymainaddress]{North Carolina State University, Raleigh, NC USA \\ email: \href{mailto:staghiy@ncsu.edu}{staghiy@ncsu.edu} }
\address[mysecondaryaddress]{Tompkins fulfillment services, Raleigh, NC USA \\ email: \href{mailto:dclengacher@gmail.com}{dclengacher@gmail.com}}
\address[mysecondaryaddress2]{North Carolina State University, Raleigh, NC USA \\ email: \href{mailto:asadegh3@ncsu.edu}{asadegh3@ncsu.edu}}
\address[mysecondaryaddress3]{North Carolina State University, Raleigh, NC USA \\ email: \href{mailto:asahebi@ncsu.edu}{asahebi@ncsu.edu}}
\address[mysecondaryaddress4]{North Carolina State University, Raleigh, NC USA \\ email: \href{mailto:rbhandfi@ncsu.edu}{rbhandfi@ncsu.edu}}

\newpageafter{author}

\begin{abstract}
Hierarchical time series demands exist in many industries and are often associated with product, time frame, or geographic aggregations. Traditionally, these hierarchies have been forecasted using “top-down”, “bottom-up”, or “middle-out” approaches. The question we aim to answer is how to utilize child-level forecasts to improve parent-level forecasts in a hierarchical supply chain. Improved forecasts can be used to considerably reduce the logistics costs, especially in e-commerce. We propose a novel multi-phase hierarchical (MPH) approach. Our method involves forecasting each series in the hierarchy independently using machine learning models, then combining all forecasts to allow a second phase model estimation at the parent level. Sales data from MonarchFx Inc. (a logistics solutions provider) is used to evaluate our approach and compare it to “bottom-up” and “top-down” methods. Our results demonstrate an 82-90\% improvement in forecast accuracy using the proposed approach. Using the proposed method, supply chain planners can derive more accurate forecasting models to exploit the benefit of multivariate data.\end{abstract}


\end{frontmatter}


\section{Introduction}\label{intro} 



The efficient movement of goods in a supply chain depends on the ability to accurately forecast product demands.  Oftentimes, these forecasts must be produced within a hierarchical structure which may represent geographic regions, product families, \citep{hyndman2011optimal}, or time periods \citep{athanasopoulos2017forecasting}.
The value of hierarchical forecasting is that it can provide decision support information to different stakeholders across various organizational functions and managerial levels \citep{fliedner1992constrained}. For instance, hierarchical forecasts can be used to improve market positioning, inventory planning, facility layouts, or increased efficiency of operational logistics and transportation networks, leading to increased customer satisfaction and lower costs. \cite{muir1979pyramid} explained how hierarchical forecasting can increase overall forecast accuracy, noting that combining data from two or more homogeneous items can produce a stabilizing effect.

Two dominant approaches exist in the hierarchical forecasting literature:  top-down and bottom-up. In the top-down approach, a forecast is initially created at an aggregated level, then disaggregated to lower levels of the hierarchy \citep{boylan2010choosing}. A common disaggregation approach involves proration \citep{fliedner1999investigation,strijbosch2007hierarchical} in which the aggregate demand forecast is multiplied by the ratio of corresponding demand to aggregate demand, resulting in an estimate for the next lower-level in the hierarchy. In the bottom-up approach, the steps are reversed. The lowest level of the hierarchy is forecasted first (i.e. SKU level), then aggregated to estimate higher levels in the hierarchy \citep{hyndman2011optimal}. A third approach called middle-out combines aspects of top-down and bottom-up. In middle-out, the forecast is performed at a middle level of the hierarchy, then aggregated up and disaggregated down to estimate the forecasts for other levels of the hierarchy.

With respect to top down forecasting, \cite{gross1990disaggregation} argued that two simple disaggregation techniques can be effective; ``average historical proportions" and ``proportions of the historical averages" \citep{athanasopoulos2009hierarchical}. In the ``average historical proportions'' approach, the share of each lower level time series of the aggregated series is calculated across all periods, i.e. a linear average share is used. In the``proportions of the historical averages'' approach, a volume weighted share across all time periods is employed.  The authors also mention that for the ``average historical proportions'' approach, one is not required to only use the historical proportions, but can utilize the forecasted proportions instead. Promising results were derived using this  approach and it is offered in some forecasting software \citep{boylan2010choosing}.

In practice, there may be multiple features in the input data (e.g. date, time, holidays, seasonal discounts, etc.) that can be leveraged to improve forecast accuracy within supply chains. To the best of our knowledge, most of the research in the supply chain hierarchical forecasting literature is univariate. We found no documented multivariate hierarchical forecasting models that employ lower level forecasts as features in parent level modeling. In this research, we employ multiple features (in contrast to univariate time series data) and child level (SKUs) and parent level (brand) forecasts in a hierarchical supply chain model to improve forecast accuracy at the parent level in the hierarchy. We utilize Machine Learning (ML) techniques including Multi-Layer Perceptron (MLP), Random Forest (RF), Gradient Boosting (GB), and Extreme Gradient Boosting (XGB) to build competing forecasting models. The rest of the paper is organized as follows: In section 2, we briefly review the various existing hierarchical forecasting methods and the aggregation approaches in use. In section 3, we present the details of our proposed Multi-Phase Hierarchical forecasting approach (MPH). We then describe numerical experiments that demonstrate the performance of MPH using sales data from MonarchFx Inc. (a logistics solutions provider) that is representative of a mid-tier supply chain customer in section 4. We summarize our conclusions and discuss practical aspects of our work in section 5.

\section{Literature Review}\label{litRev}
The performance of top-down and bottom-up forecasting approaches in the literature are mixed \citep{syntetos2016supply}. Some authors found top-down approaches to be superior \citep{barnea1980analysis,gross1990disaggregation,fliedner1999investigation}, while others found bottom-up methods to be more accurate. \citep{dangerfield1992top,gordon1997top}. These conflicting results occur because the performance of each approach depends on the nature of demand for the products involved. To illustrate, consider a three-level product hierarchy, with product sales at the lowest level, group sales at the middle level, and category sales at the top level. Since group sales are determined by the sum of product sales (given the additive nature of the hierarchy), and the sum of group sales determines category sales, the underlying demand process is transformed at different levels of the hierarchy. When aggregating, a significant loss of information can occur, which tends to render bottom-up forecasting more favorable. Conversely, in the top-down approach, benefits can occur due to random noise cancellation \citep{fliedner1999investigation}. Because the performance of each approach depends on the demand generation process within the data, a wide range of conflicting results appears in the literature. Thus, depending on the demand process and parameter settings, one approach may perform better than the other in different contexts \citep{widiarta2007effectiveness,widiarta2009forecasting}.

An early study comparing both top-down and bottom-up approaches was conducted by \cite{grunfeld1960aggregation}, in which they found the top-down approach more accurate, with the explanation that disaggregated data is more susceptible to error. \cite{fogarty1983production} and \cite{narasimhan1995production} derived similar conclusions in their work. Conversely, the loss of information in a top-down approach was considered substantial in \cite{orcutt1968data} and leading to the conclusion that the bottom-up approach is superior. In \cite{shlifer1979aggregation}, the authors identified conditions on the hierarchy's structure and forecast horizon, under which they concluded that the bottom-up approach is favorable. The robustness and bias of both approaches were investigated in \cite{schwarzkopf1988top}. The authors concluded that the bottom-up approach is more favorable unless there exist unreliable or missing data at the bottom of the hierarchy.

A significant characteristic of the underlying demand process involves the dependencies between demand produced at each level, which can be a reason for the performance differences between top-down and bottom-up approaches \citep{chen2007use}. 

\cite{sbrana2013forecasting} summarizes the arguments that are often made against top-down approaches. First, he states that the a high (or low) variance in one level in a hierarchy may be indicative of high (or low) variance at other levels. In such cases, allocating measures of variance from higher levels to lower levels in a hierarchy may yield better results. Second, since different products may be classified in different segments, the aggregation of data will lead to a loss of information, making the top-down approach less appealing.  

On the other hand, there are examples in the supply chain forecasting literature where the authors favor the top-down approach. In \cite{boylan2010choosing}, the author found that aggregated data can lead to more accurate sales forecasts when dealing with change policies (e.g. change in pack sizes), compared to individual level forecasts. In such cases, common disaggregation techniques (``average historical proportions'' and ``proportions of the historical averages'') may not be useful, and judgmental estimates are required in disaggregation methods to handle such changes in policy. 

One method to overcome these drawbacks involves analysis of the conditions in which each approach produces superior forecasting accuracy outcomes. In \cite{widiarta2008forecasting}, the top-down and bottom-up approaches are compared in the context of production planning. Their goal was to estimate requirements at the SKU level. The aggregate demand series were assumed to have correlated sub-aggregate components, each of which were assumed to follow a first order univariate moving average (stationary) process correlated over time. They concluded that both methods have nearly identical performances. Later, \cite{widiarta2009forecasting} investigated the relative effectiveness of bottom-up and top-down approaches to forecast demand at the aggregate level rather than the SKU level. They concluded that when all sub-aggregate components of the time series follow a first-order univariate moving average process with identical coefficients of the serial correlation term, the relative performance of both top-down and bottom-up approaches are similar. Additionally, the different coefficients of the serial correlation term among sub-aggregate components were examined in a simulation study. The result was that the differences in the performance are relatively insignificant when there are small or moderate correlations between the sub-aggregate components. \cite{sbrana2013forecasting} found that when moving average parameters are not identical, the performance of top-down and bottom-up approaches is similar. 

More recently, \cite{rostami2015non} analyzed theoretically and by means of simulation (using theoretically generated data) the relative performance of top-down and bottom-up forecasting methods for both aggregate and SKU level demand. The latter was assumed to follow a non-stationary ARIMA (0,1,1) demand process and exponential smoothing (which is optimal for this demand process). An important finding was that the forecast accuracy improvements achieved by bottom-up and top-down methods for non-stationary demands are higher than those associated with stationary cases. The theoretical findings were validated through empirical analysis on data from a European superstore. 

A limitation observed in this work is that the generation of forecasts is dominated by the time series at a single level of aggregation (the point at which forecasts are created). To overcome this issue, a regression-based approach was introduced by \cite{hyndman2011optimal}. In their approach, they estimated the time series at multiple hierarchy levels and then optimized this combination using linear regression. This approach sought to derive the benefits of an ensemble of bottom-up and top-down approaches, employing a linear combination of both. Their method demonstrated a significant improvement in forecast accuracy compared to the traditional approaches. This improvement was believed to be a function of employing a combination of forecasts that reduced the variance of forecast error \citep{timmermann2006forecast,barrow2016distributions}. \cite{hyndman2011optimal} conclude that their proposed combination method is “optimal”, and compared to all combination forecasts, leads to the least variance. Their work is inspired by earlier research in economics focusing on revising measurements of macro-economic indicators \citep{zellner2000note,espasa2002forecasting,hubrich2005forecasting}. Other research focuses on using different sources to combine forecasts, e.g. utilizing different available information provided by human experts \citep{budescu2014identifying,lamberson2012optimal}. Additionally, the combination of forecasts may reduce model specification and estimation uncertainty \citep{kourentzes2014neural}. In a later work, \cite{hyndman2014optimally} demonstrate the extendibility of their combination approach for hierarchical forecasting to non-hierarchical time series, and time series with partial hierarchical structure. They also proposed a solution to solve the scalability problem that existed in their previous paper \cite{hyndman2011optimal}. They use a linear model structure for a more efficient coefficient estimation.

In \cite{pennings2017integrated}, the authors utilize all the series in a hierarchy in contrast to a top-down or bottom-up approach. They then incorporate a Kalman filter and state space model to comprehend the dependencies between products (e.g. product substitution, product complementarity). Using a multi-variate state space, one is able to estimate the hierarchical time series efficiently using a Kalman filter as a prediction error decomposition tool \citep{durbin2012time}. In this manner, multiple methods for forecasting hierarchical time series exist \citep{hyndman2008automatic,snyder2012forecasting}. In their approach, forecasts for the aggregate level is derived by summing the forecast of product sales at the base level. The Kalman filter is then used to track the forecast error of individual series at each level of the hierarchy back to the associated states. In this manner, the forecast leverages the information from all series. The authors conclude that their approach is superior to the traditional top-down and bottom-up approach since they incorporate information from all levels of the hierarchy.

Our work builds on the research by \cite{hyndman2011optimal}, and \cite{pennings2017integrated} (discussed previously), in which they combine information at all levels of hierarchy to improve forecasting accuracy. However, these authors only employ univariate data as their input, and do not leverage multiple features. Our main contribution in this paper is to propose a novel approach which 1) utilizes forecasts at lower levels to improve forecasts at higher levels, 2) uses multivariate data at each level of the hierarchy instead of univariate data, which is more commonly seen in the literature, and 3) leverages machine learning models.  The latter component is, to the best of our knowledge, a novel application in the supply chain forecasting literature. To achieve our goal, we propose an MPH approach which is discussed in the following section.

\section{Multi-Phase Hierarchical Forecasting Approach}
Our goal is to find a small loss value, $l(.)$, in the parent level of the hierarchy, to optimize:
\begin{equation}
    \min_\omega \frac{1}{n} \sum_{i=1}^{n}l(\theta(x_i;\omega),y_i)
\end{equation}

where $\omega$ is the matrix of the weights, $x_i$ is the vector of the inputs from the $i^{th}$ instance, $y_i$  is the dependent variable, e.g., demand (sales) values, and $\theta(.)$ is the output function defined by the forecasting model. The well-known Mean Absolute Error (MAE) is being used as our loss function, $l(.)$, in which, the average of differences between the actual demands and estimated demand is calculated.

To achieve a higher level of accuracy in the parent level of the hierarchy, we utilize an MPH approach. In the first phase, we forecast at both child level (SKU level) and parent level (brand) demands using several machine learning approaches. Then, for each individual time series, we select the most promising forecast method in terms of MAE. MAE is calculated based on a cross-validation technique. In the second phase, we aggregate the forecasts at the child level and parent level and use them as an input for the multi-feature forecasting approach.

\subsection{Overview of Forecasting Methods}
Conventional parametric forecasting techniques include ARIMA, GARCH, and TRANSFER models \citep{box2015time,shumway2011time}. Moreover, \cite{taylor2000quantile} forecasts the demand for time steps ahead using a normal distribution. However, in the situations where demand values are volatile and correlated over time, their model does not yield good performance. One way to overcome this issue is to use a class of algorithms called “universal approximators”. This class of algorithms is based on machine learning techniques and is able to approximate any function given an arbitrary forecast accuracy. These approximators can learn any function of past and future data and therefore other forecasting models can be considered as a subset of the functions which they are able to learn. Machine Learning (ML) techniques, such as Multi-Layer perceptron (MLP), Random Forest (RF), Gradient Boosting (GB), and eXtreme Gradient Boosing (XGB) are some of these universal approximators, which are able to be used to learn any function and have many applications in practice \citep{belgiu2016random,mei2014random,rahmati2016application,de2007boosted,moisen2006predicting,chen2016xgboost,cigizoglu2004estimation,hippert2001neural,deo2018multi}.

Supply chain forecasting is a field which generally consists of very “noisy” data, thus it is important to control for noise and learn the true underlying demand patterns which are likely to be repeated in the future. The universal approximators discussed earlier have two desirable features which make them suitable for the supply chain forecasting problem, while dealing with noise. The first is that they are capability of learning any arbitrary function, while the second feature is the capability to control the learning process.  

Since we want to exploit additional information provided by multiple input features, we are faced with a multi-dimensional input data vector. The traditional parametric forecasting models such as ARIMA are not able to integrate multi-dimensional inputs, thus we exploit the ability of universal approximators to take multi-dimensional inputs and utilize them in our forecasting model. The details of the MPH approach are explained in the following subsection.

\subsection{MPH Algorithm}
Our proposed approach, the multi-phase hierarchical method, is a novel method that utilizes machine learning models to independently forecast each series in a hierarchy. This means that for each level in the hierarchy, a separate machine learning model is trained to forecast the time series data for that level. Once the forecasts for each level have been generated, they are then combined in a second phase of the method, where a model estimation is performed at the parent level. This two-phase approach allows for a more accurate and robust forecasting method, as the forecasts from each level are combined and considered at the parent level. This process is repeated at each level of the hierarchy, allowing for a comprehensive and detailed forecasting method. Algorithm~\ref{alg:framework} shows the steps in the proposed MPH method.

\begin{algorithm}[hbt!]
\caption{MPH Algorithm}\label{alg:framework}
\begin{algorithmic}
\STATE (Phase I)
\STATE \textbf{Input:} 
\bindent
\STATE Choose forecasting model types which support multi-feature inputs, e.g. MLP, RF, GB, XGB, etc. ,
\STATE Set $i=1$, 
\STATE Suppose we have chosen $N$ forecasting approaches. 
\eindent
\STATE \textbf{Output:}  Best forecasting output.
\STATE \textbf{Step 1.} Use the $i^{th}$ forecasting approach to forecasting parent-level demand (brand demand) and child-level demands (SKU demand).
\STATE \textbf{Step 2.} Optimize the hyperparameters of the $i^{th}$ forecasting method using a search approach, e.g. Bayesian optimization method, grid search, successive halving.
\STATE \textbf{Step 3.} $i \gets i+1$
 \bindent
\IF{$i=N+1$}
    \STATE go to \textbf{Step 4.}
\ELSIF{$i \le N$}
    \STATE go to \textbf{Step 1.}
\ENDIF
\eindent
\STATE \textbf{Step 4.} Using the outputs of the previous steps, record the best forecasting approach and the associated outputs for demands at all levels.
\\
\STATE (Phase II)

\STATE \textbf{Step 5.} Append the recorded outputs of \textbf{Step 4} to input data of the parent level, as additional features.
\STATE \textbf{Step 6.} Repeat \textbf{Steps 1 and 2} once more, using the new input data with additional features. The only modification is to only forecast the parent level. 
\STATE \textbf{Step 7.} Choose the best forecasting output among the forecasting methods in \textbf{Step 6}.

\end{algorithmic}
\end{algorithm}

\begin{figure}[htbp]
        \center{\includegraphics[width=\textwidth]
        {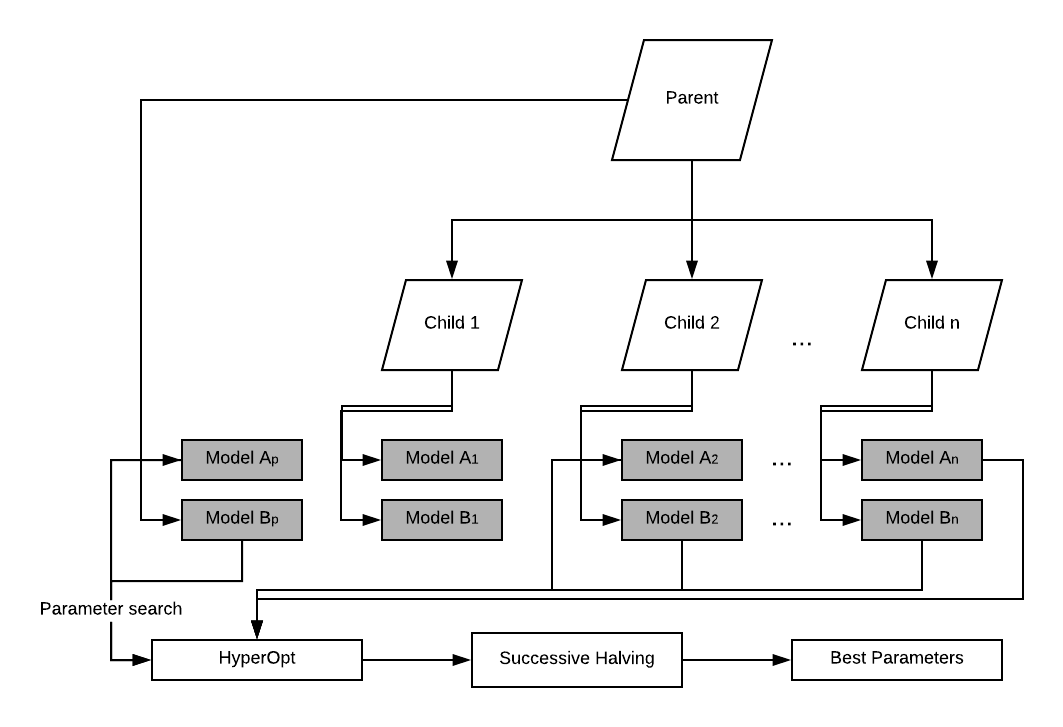}}
        \caption{\label{fig:PhaseIdiagram} Phase $I$ of MPH forecasting model}
\end{figure}
      
Figure \ref{fig:PhaseIdiagram} provides an overview of Phase $I$ for this procedure, in which two forecasting models were chosen as base forecasting approaches. Model A can be a tree-based forecasting model, e.g. RF, GB, or XGB, and model B can represent an exploration-capable model, e.g. artificial neural network models such as MLP. As depicted in figure \ref{fig:PhaseIdiagram}, we have a two-level hierarchical structure with 1 parent and n children. In Phase $I$ of the model, we use the selected models (i.e. models A and B) to forecast demands at both parent and child levels. Since we are dealing with universal approximators, they have several hyperparameters, on which the model is very sensitive in term of accuracy. Hence, one needs to find an approach to optimize the hyperparameters of the forecasting models.

\subsection{Hyperparameter Optimization}
There are several approaches in the literature that address hyperparameter optimization in machine learning \citep{maclaurin2015gradient,feurer2015initializing,li2017hyperband,bergstra2013hyperopt}. In this paper, we use the hyperOpt algorithm proposed by \cite{bergstra2013hyperopt}, and combine it with the successive halving approach \citep{jamieson2016non} to obtain a more efficient search. In the following a summary of the HyperOpt algorithm is provided and the details of the proposed hyperparameter optimization algorithm is explained.

\subsubsection{HyperOpt:}
HyperOpt is a module proposed by \cite{bergstra2013hyperopt} , and is focused on intelligently searching through the hyperparameter space. One approach is to use the Tree-structured Parzen Estimator (TPE) algorithm \citep{bergstra2011algorithms}, in which the search space is explored in an intelligent way, while the parameter values are narrowed down to the best estimated parameters. In contrary to the Grid Search, in which the hyperparameters must be pre-determined and the increment steps fixed, HyperOpt is an oriented random search and is proven to work efficiently \citep{bergstra2013hyperopt}. Hence, it serves as a good candidate to tune and optimize hyperparameters for universal approximators, as adopted in this paper.

\subsubsection{Proposed Hyperparameter Optimization Algorithm:}
A hyperparameter optimization algorithm is a technique used to select the best set of hyperparameters for a machine learning model. Hyperparameters are parameters that are not learned from the data during training, but are set by the user before training. The goal of a hyperparameter optimization algorithm is to find the best combination of hyperparameters that results in the highest performance for the model on a validation dataset. In this section, we propose another algorithm for screening procedure in the first phase and we add the best-performing forecasts as an additional feature:
\begin{itemize}
    \item \textit{Initializing the sample space by HyperOpt.} Suppose that we choose to start with $N$ parameter settings to search more rigorously among them. In our modified approach, we use HyperOpt to search intelligently through the search space, and we store the first $N$ parameter settings that are used by HyperOpt. Note that each HyperOpt iteration is only performed on one set of train/test data. Now we use the generated $N$ parameter settings as an input for a more rigorous search by successive halving \citep{jamieson2016non}. 
    \item \textit{We follow the idea of successive halving proposed by \cite{jamieson2016non}}. Using $N$ parameter settings generated by HyperOpt, the well-known K-fold algorithm \citep{kohavi1995study} is used to evaluate each parameter settings for a fixed amount of time/budget, e.g. T. Then, we select the top-performing half of the parameter settings $(N/2)$, and again, we evaluate them via k-fold for time 2T. This procedure is repeated until the search space is singular or the designated budget is exhausted. 
\end{itemize}

\begin{figure}[htbp]
        \center{\includegraphics[width=\textwidth]
        {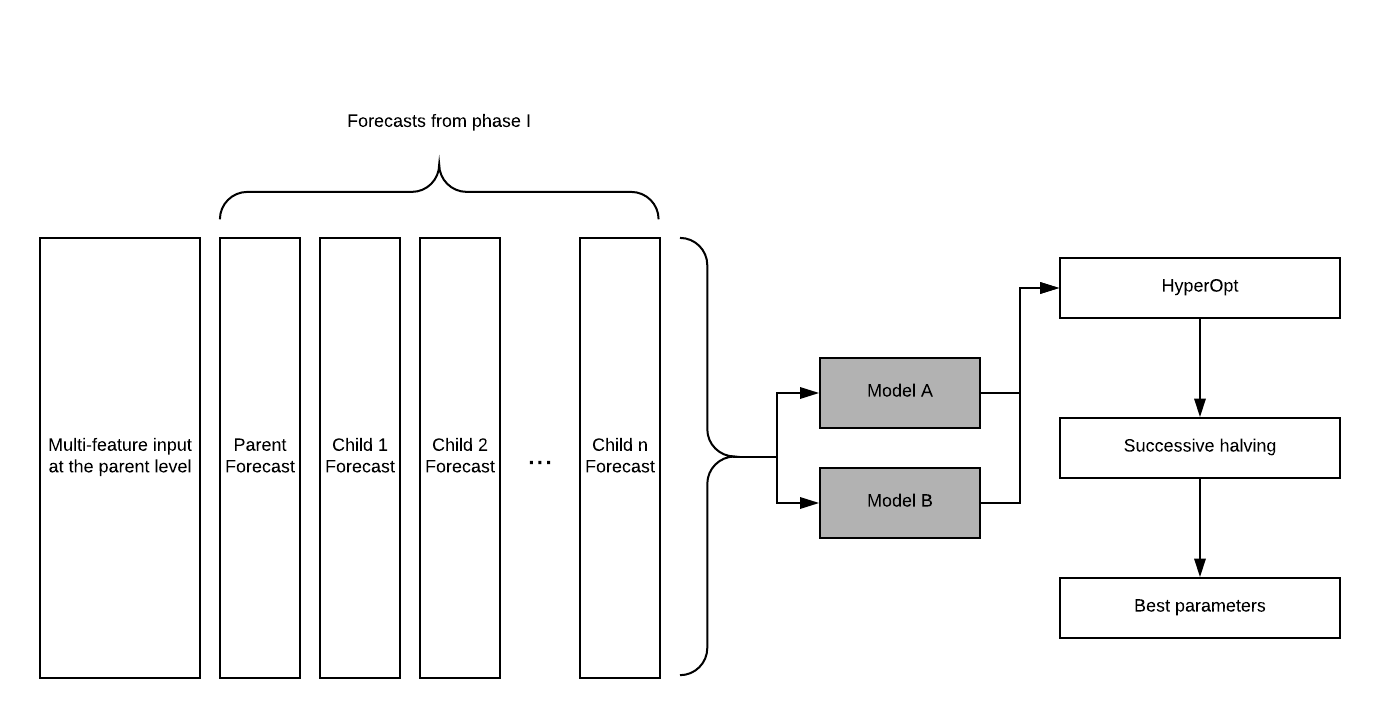}}
        \caption{\label{fig:PhaseIIdiagram} Phase $II$ of MPH forecasting model}
      \end{figure}
      
The idea behind the above algorithm is quite intuitive. Initially, HyperOpt is used as the screening procedure on the search region by expending a small budget of processing time. After initial candidates (parameter settings) were selected, successive halving is utilized for a more rigorous evaluation. This procedure spends computational budget more efficiently by focusing on the parts of search region which have more potential.
      
In the second phase we add the best performing forecasts as additional features to the input data of the parent (See Figure \ref{fig:PhaseIIdiagram}). Next, a parent level forecasting model is re-estimated  using the new input and then the hyperparameter optimization process is conducted. After identifying the best parameter settings, we select the best performing forecast as our final forecasting model. 

\section{Numerical Experiment}
The MPH forecasting algorithm was implemented on sales data provided by MonarchFx Inc., which consists of 935 days of data for ten Stock Keeping Units (SKUs) and aggregated data which represents total brand sales. This data is representative of one of MonarchFx's mid-tier supply chain customers. In addition to the historical sales data, the input also contained additional features including:
\begin{itemize}
    \item Promotion: a binary variable indicating if a promotion was present.
    \item Holiday: a binary variable indicating holiday periods.
    \item Day of the week: seven dummy variables corresponding to day of the week.
    \item Date: in the format day/month/year
\end{itemize}

Each of these factors may increase the predictive power of forecasting models, both independently as well as in combinations. 

To measure model accuracy, the Mean Absolute Error (MAE) is used:
\begin{equation}
    MAE=\sum_i\mid y_i-y^{\hat{}}_i\mid
\end{equation}

Where $y_i$ corresponds to the actual values of sales, and $y_i^{\hat{}}$ is the forecasted sales on day $i$.

The well-known k-fold cross-validation method \citep{kohavi1995study} with k=5 is used to test each forecasting model and measure the forecast accuracy. The parameter k refers to the number of groups that the input data will be divided to. We chose this method because it provides a less biased estimate of the model compared to a single train/test split of the data. In this procedure, initially the data is randomly shuffled and is divided into k different groups. Then, for each group, it is selected as the test dataset and the remaining data is considered as the train set. The forecasting model is trained on the train set and the accuracy is measured on the test set. This procedure is repeated for every group and the average MAE across k train/test splits is reported as the final MAE.

Multi-Layer Perceptron (MLP), Random Forest (RF), Gradient Boosting (GB), and Extreme Gradient Boosting (XGB) are the forecasting models selected for the experiments, due to their popularity in machine learning forecasting literature \citep{mei2014random,de2007boosted,cigizoglu2004estimation,deo2018multi,zikeba2016ensemble}. Using these four forecasting models, the MPH algorithm in conjunction with the hyperparameter optimization method (explained in section 3-2) is implemented on the data and the forecasting error at the parent level is compared to the top-down and bottom-up approach.

Tables \ref{table:PhaseISKU} and \ref{table:PhaseIParent} show the results of Phase $I$ of algorithm for the lower level and parent level of the hierarchy, respectively. Table \ref{table:PhaseISKU} contains 10 rows corresponding to each SKU. MAE is reported for each of four forecasting methods (after performing k-fold cross-validation), and the forecasting method with the minimum MAE is selected for phase $II$.

\begin{table}[]
\centering \caption{Phase $I$ child-level results } \label{table:PhaseISKU}
\begin{tabular}{llllllll}
\hline
Series & MLP & RF  & GB  & XGB & Best & Range & Min MAE \\ \hline
1      & 370 & 339 & 366 & 350 & RF   & 31    & 339     \\ 
2      & 404 & 381 & 405 & 388 & RF   & 24    & 381     \\ 
3      & 607 & 557 & 609 & 588 & RF   & 52    & 557     \\ 
4      & 681 & 684 & 725 & 708 & MLP  & 44    & 681     \\ 
5      & 364 & 343 & 389 & 360 & RF   & 46    & 343     \\ 
6      & 408 & 385 & 397 & 405 & RF   & 23    & 385     \\
7      & 676 & 691 & 732 & 709 & MLP  & 56    & 676     \\ 
8      & 446 & 421 & 449 & 451 & RF   & 30    & 421     \\
9      & 537 & 537 & 550 & 537 & MLP  & 13    & 537     \\ 
10     & 395 & 363 & 385 & 375 & RF   & 32    & 363     \\ \hline
\end{tabular}
\end{table}

\begin{table}[]
\centering \caption{Phase $I$ Parent level results} \label{table:PhaseIParent}
\begin{tabular}{lllllll}
\hline
MLP  & RF   & GB   & XGB  & Best & Range & Min MAE \\ \hline
3972 & 3182 & 3068 & 3118 & GB   & 904   & 3068    \\ \hline
\end{tabular}
\end{table}

The results of phase $I$ are added as additional features to input data at the parent level. As phase $II$ of the algorithm suggests, MLP, RF, GB, and XGB models are estimated again using the new input data. Table \ref{table:PhaseII} reports the MAE (after performing k-fold cross-validation) at the end of phase $II$ for each of the forecasting models. The minimum MAE is selected as the final MAE at the parent level, which is 303. 

\begin{table}[]
\centering \caption{Phase $II$ results} \label{table:PhaseII}
\begin{tabular}{lllllll}
\hline
MLP & RF  & GB  & XGB & Best & Range & Min MAE \\ \hline
445 & 610 & 528 & 303 & XGB  & 307   & 303     \\ \hline
\end{tabular}
\end{table}

The final MAE of MPH algorithm is compared to the MAE of top-down and bottom-up approach, in table \ref{table:Comp_td_bu}.

\begin{table}[]
\centering \caption{Top-down, Bottom-up vs MPH MAE} \label{table:Comp_td_bu}
\begin{tabular}{lll}
\hline
MPH & Hierarchy & \% of Improvement \\ \hline
303 & Top-down: 3068 &  90\%              \\ 
&  Bottom-up: 1672 & 82\%              \\ \hline

\end{tabular}
\end{table}

As the final MAE results suggest (Table \ref{table:Comp_td_bu}), comparing MPH algorithm to both top-down and bottom-up approached, 90\% and 82\% improvement is gained, respectively. These outcomes demonstrate the advantages of MPH in substantially improving forecasting accuracy. The reason lies in the fact that the information at the child level is leveraged to improve forecasting accuracy at the parent level, which was previously ignored in both top-down and bottom-up approaches.

To show the accuracy improvement we can get by using MPH algorithm, we compare our results to output of machine learning models that we used as the basis of MPH. The results are shown in table \ref{table:PhaseIIToML}. As the results suggest, we are gaining at least 90\% improvement in forecast accuracy over popular machine learning models, which shows the significant improvement in the results obtained from MPH.

\begin{table}[]
\centering \caption{Comparing results of MPH to forecasts from machine learning models  } \label{table:PhaseIIToML}
\begin{tabular}{llll}
\hline
Machine learning model & MAE & MAE from MPH& Improvement\\ \hline
MLP      & 3972 & 303 & 92\%  \\ 
RF      & 3182 & 303 & 90\%      \\ 
GB      & 3068 & 303 & 90\%\\ 
XGB     & 3118 & 303 & 90\%     \\ \hline

\end{tabular}
\end{table}

For the sake of completeness, we also compare the results of our algorithm to traditional time series forecasting methods, namely nai\"ve forecasting, moving average, simple exponential smoothing, Holt's linear trend, Holt-Winter's additive method, ARIMA, theta and ARIMAX. Tables \ref{table:PhaseIToTraditional} and \ref{table:PhaseIIToTraditional} show the results of comparing the aforementioned time series forecasting methods' results with phase I and Phase II output of MPH algorithm.

\begin{table}[]
\centering \caption{Comparing phase $I$ results of MPH to traditional time series forecasting methods  } \label{table:PhaseIToTraditional}
\begin{tabular}{llll}
\hline
Forecasting method & MAE & MAE from MPH (Phase $I$)  & Improvement\\ \hline
Nai\"ve forecasting      & 24974 & 3068 & 88\%  \\
Moving average      & 20647 & 3068 & 85\%      \\ 
Simple exponential smoothing      & 10120 & 3068 & 70\%\\ 
Holt's linear trend     & 18681 & 3068 & 84\%     \\ 
Holt-Winter's additive method      & 12076 & 3068 & 75\% \\ 
ARIMA      & 3979 & 3068 & 23\%\\ 
Theta      & 19743 & 3068 & 84\%\\ 
ARIMAX      & 3364 & 3068 & 9\%\\ \hline

\end{tabular}
\end{table}

\begin{table}[]
\centering \caption{Comparing phase $II$ results of MPH to traditional time series forecasting methods  } \label{table:PhaseIIToTraditional}
\begin{tabular}{llll}
\hline
Forecasting method & MAE & MAE from MPH (Phase $II$)  & Improvement\\ \hline
Nai\"ve forecasting      & 24974 & 303 & 99\%  \\ 
Moving average      & 20647 & 303 & 99\%      \\ 
Simple exponential smoothing      & 10120 & 303 & 97\%\\ 
Holt's linear trend     & 18681 & 303 & 98\%     \\ 
Holt-Winter's additive method      & 12076 & 303 & 97\% \\ 
ARIMA      & 3979 & 303 & 92\%\\ 
Theta      & 19743 & 303 & 98\%\\ 
ARIMAX      & 3364 & 303 & 91\%\\ \hline

\end{tabular}
\end{table}

\begin{figure}[htbp]
        \center{\includegraphics[width=\textwidth]
        {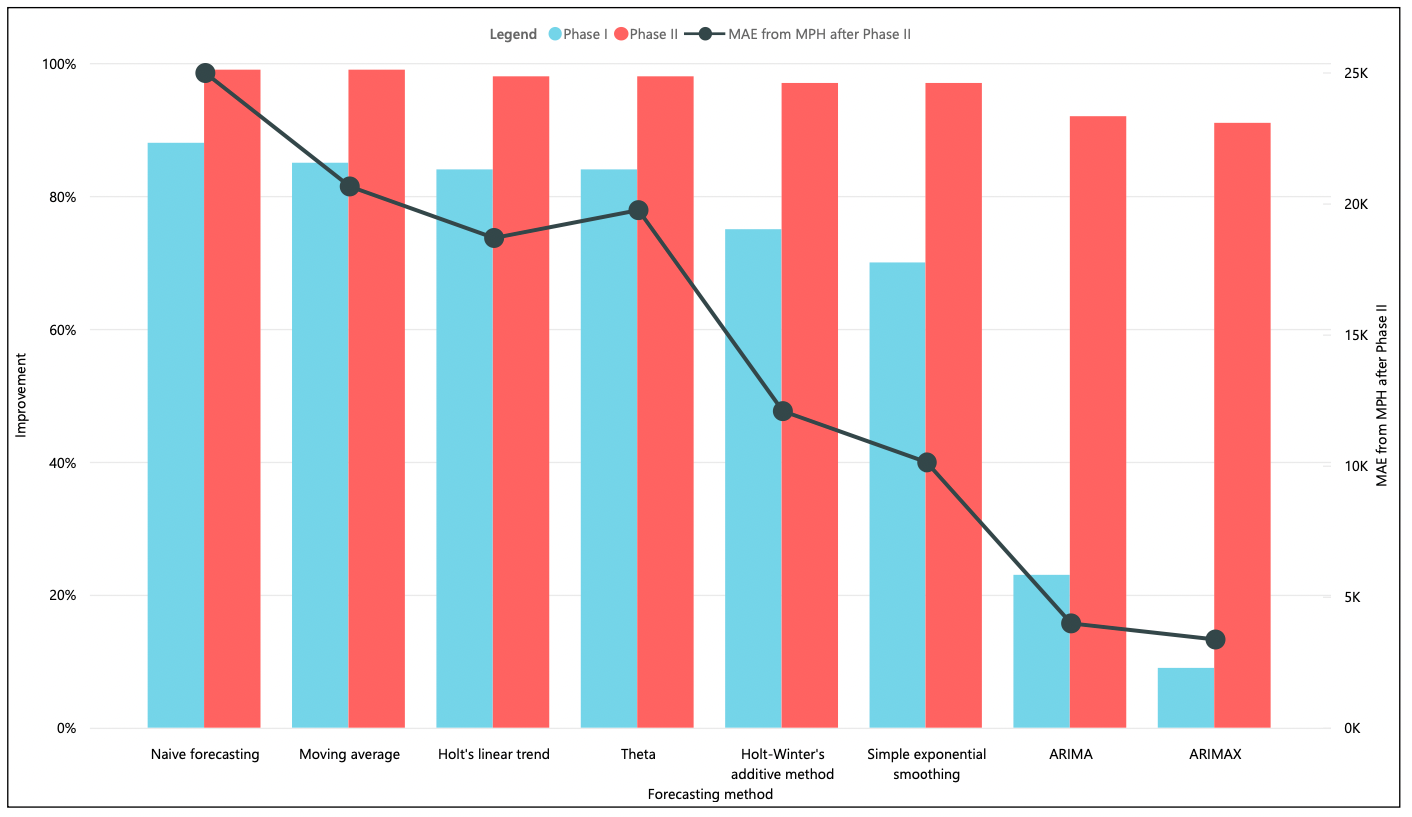}}
        \caption{\label{fig:performance} Performance of MPH algorithm under different ML forcasting models.}
\end{figure}

As we can from the results of tables \ref{table:PhaseIToTraditional} and \ref{table:PhaseIIToTraditional}, MPH algorithm performs significantly better than traditional time series forecasting methods. The main reason behind this significant improvement is twofold. First, in contrast to traditional forecasting methods, which mostly use univariate time series, we use multiple features as input variables in MPH algorithm. The second reason is that MPH algorithm uses information at both levels of the hierarchy (SKU level and brand level), which helps the algorithm to provide significantly more accurate forecasts. Figure~\ref{fig:performance} summarizes the improvement using the MPH algorithm for different ML models.

\section{Conclusions}
In this paper, we develop a novel multi-phase hierarchical approach (MPH) for supply chain forecasting using machine learning techniques supporting multi-feature input data (e.g. MLP, RF, GB, and XGB). In the proposed two-phase model, the information at the child level is leveraged to improve forecasting accuracy at the parent level, by adding the results of the best forecasting model for each child as additional features at the parent level. The MPH algorithm is implemented on sales data provided by MonarchFx Inc. and the results were compared to a top-down and bottom-up approach. The results demonstrate that a considerable improvement can be achieved by utilizing the MPH algorithm (90\% improvement in comparison with top-down approach, and 82\% improvement comparing to bottom-up approach). This improvement is possible due to the fact that the MPH algorithm leverages information both at the child level and parent level.

Based on the experience of one of the co-authors who leads the supply chain analytics function at MonarchFx Inc., there are multiple applications possible for our approach. Indeed, the majority of companies employing supply chain forecasting solutions generally apply top-down and bottom-up approaches and use traditional models that only support single feature input data. However, in practice, multiple factors can impact future sales and can be controlled for in this manner to improve forecast accuracy. Using the machine learning forecasting models developed in this paper, supply chain planners can derive more accurate forecasting models to exploit the benefit of multivariate data. 

There are multiple possible future extensions to this work. One is to use the MPH algorithm on hierarchies with more than two levels. The other is to utilize the reconciliation techniques used by \cite{hyndman2011optimal}. Another possible path is to characterize the situations under which each of the forecasting models perform best in the child level and parent level. Researchers may also use the forecasting model selection method developed by \cite{taghiyeh2020forecasting} to select the best forecasting model among existing machine learning models. The noisy optimization method in \cite{taghiyeh2016new} may also be used to find the parameters for optimal reconciliation for the levels of the hierarchy. To improve the speed of the model, the parallelization method proposed in \cite{rosen2016parallel} can be utilized. We believe there lies great promise for using these approaches in the future.

\vspace{6pt} 

\textbf{Funding:}
This research received no external funding.

\textbf{Institutional Review Board Statement:}
Ethical review and approval are Not applicable (Due to studies not involving humans or animals.)

\textbf{Informed Consent Statement::}
Not applicable.

\textbf{Acknowledgments:}
We thank our colleagues from Supply Chain Resource Cooperative (SCRC) at North Carolina State University and MonarchFx Inc. who provided insight and expertise that greatly assisted the research, although they may not agree with all of the interpretations/conclusions of this paper.

\textbf{Conflicts of Interest:}
The authors declare no conflict of interest.

\textbf{Abbreviations}
The following abbreviations are used in this manuscript:\\
\noindent 
\begin{tabular}{@{}ll}
MPH & multi-phase hierarchical\\
SKU & Stock keeping unit\\
ML & Machine Learning \\
MLP & Multi-Layer Perceptron\\
RF & Random Forest\\
GB & Gradient Boosting\\
XGB & Extreme Gradient Boosting\\
MPH & Multi-Phase Hierarchical forecasting approach\\
MAE & Mean Absolute Error\\
TPE & Tree-structured Parzen Estimator
\end{tabular}

\section*{References}
\bibliography{MHP}

\end{document}